\documentclass{aims}
\usepackage{cite}
\usepackage{multirow}
\usepackage{graphicx}
\usepackage{txfonts}
\def\typeofarticle{Research article}
\def\currentvolume{19}
\def\currentissue{x}
\def\currentyear{2023}

\def\ppages{xxx--xxx}
\def\DOI{}
\def\Received{}
\def\Revised{}
\def\Accepted{}
\def\Published{}

\numberwithin{equation}{section}

\begin{document}

\title{SegT: Separated Edge-guidance Transformer Network for Polyp Segmentation}

\author{%
  Feiyu Chen\affil{1},
  Haiping Ma\affil{1}
  and
  Weijia Zhang\affil{1,2}\corrauth
}

% \shortauthors is used in copyright information in the end of the paper
\shortauthors{the Author(s)}

\address{%
  \addr{\affilnum{1}}{Department of Mathematics, Physics and Information Sciences, Shaoxing University, Shaoxing, Zhejiang 312000, China}
  \addr{\affilnum{2}}{Department of AOP Physics, Visiting Scholar, University of Oxford, Oxford OX1 3PU, United Kingdom}}

% corresponding author
\corraddr{zhangusx@yeah.net
}

\begin{abstract}
Accurate segmentation of colonoscopic polyps is considered a fundamental step in medical image analysis and surgical interventions. Many recent studies have made improvements based on the encoder-decoder framework, which can effectively segment diverse polyps. Such improvements mainly aim to enhance local features by using global features and applying attention methods. However, relying only on the global information of the final encoder block can result in losing local regional features in the intermediate layer. In addition, determining the edges between benign regions and polyps could be a challenging task. 
To address the aforementioned issues, we propose a novel separated edge-guidance transformer (SegT) network that aims to build an effective polyp segmentation model. A transformer encoder that learns a more robust representation than existing CNN-based approaches was specifically applied. To determine the precise segmentation of polyps, we utilize a separated edge-guidance module consisting of separator and edge-guidance blocks. The separator block is a two-stream operator to highlight edges between the background and foreground, whereas the edge-guidance block lies behind both streams to strengthen the understanding of the edge. Lastly, an innovative cascade fusion module was used and fused the refined multi-level features. To evaluate the effectiveness of SegT, we conducted experiments with five challenging public datasets, and the proposed model achieved state-of-the-art performance. 

\end{abstract}


\maketitle

\section{Introduction}

According to the reports published by Globocan’2020, colorectal cancer (CRC) is the second most prevalent cancer type worldwide in terms of mortality and the third most pervasive disease across the globe\cite{sung2021global}. Colorectal polyps are abnormal tissue growths in the lining of the colon that are a precursor to CRC. After 10 to 15 years, polyps can turn into cancer if they are not treated. The best way to lower the prevalence of CRC is by early detection and effective treatment. Colonoscopy is the gold standard method of examining the gastrointestinal tract. It is used to find polyps and remove them before they turn into cancer. However, colonoscopy is a highly operator-dependent procedure, and one in four polyps may be missed during a single colonoscopy owing to human factors, such as clinician skill or subjectivity\cite{ahn2012miss}. In addition, there is evidence that absence or incomplete resection of the tumor are two key factors in the development of cancer after colonoscopy\cite{le2014postcolonoscopy}. Therefore, an automatic and accurate polyp segmentation method is needed to help doctors locate.

Despite significant progress in deep learning \cite{lecun2015deep, he2016deep}, automatically segmenting polyps remains a formidable challenge. Polyps, which result from abnormal cell growth in the human colon, are strongly related to their surrounding environment. They can vary in shape, size, texture, and color, making their appearance highly diverse.
One of the significant difficulties in polyp segmentation arises from the fact that the edges of polyps and the surrounding mucosa are not always clearly distinguishable during colonoscopy. This ambiguity is particularly pronounced in different lighting conditions and when dealing with flat lesions or inadequate bowel preparations. Consequently, the learning model for polyp segmentation faces considerable uncertainty due to these factors.
In summary, despite advancements in machine learning and computer vision, the automatic segmentation of polyps remains challenging due to the wide variety of polyp appearances and the difficulties associated with accurately identifying polyp edges and mucosal boundaries in colonoscopy images. These factors introduce significant uncertainty into the learning process, making the task particularly demanding.

In recent years, the rapid development of deep learning has led to an increasing number of deep convolutional neural networks (DCNNs) \cite{brandao2018towards, shen2023triplet, fan2020pranet, ronneberger2015u, zhang2020adaptive, zhou2018unet++} being proposed for polyp image segmentation. Brandao et al. \cite{brandao2018towards} introduced the fully connected convolution network (FCN) into the polyp region extraction issue by converting AlexNet, VGG, and ResNets into FCNs. U-shaped \cite{ronneberger2015u,zhou2018unet++} architecture containing an encoder and a decoder built up from convolutional layers is widely used for segmentation tasks with impressive performance. However, the limitation of convolutional neural networks (CNNs) indicates that the receptive field is limited, and the model only obtains local information but disregards spatial context and global information. In addition, CNNs behave similarly to a series of high-pass filters, favoring high-frequency information. Transformer \cite{carion2020end, dosovitskiy2020image, shen2023git, shen2023pbsl, vaswani2017attention} is a recently proposed deep neural network architecture. Compared with CNN, the self-attention layer in transformer is similar to a low-pass filter and can effectively identify long-term dependencies. Therefore, combining the advantages of convolutional and self-attention layers can improve the representation ability of deep networks.

\begin{figure}[t]
\begin{center}
\includegraphics[scale=0.5]{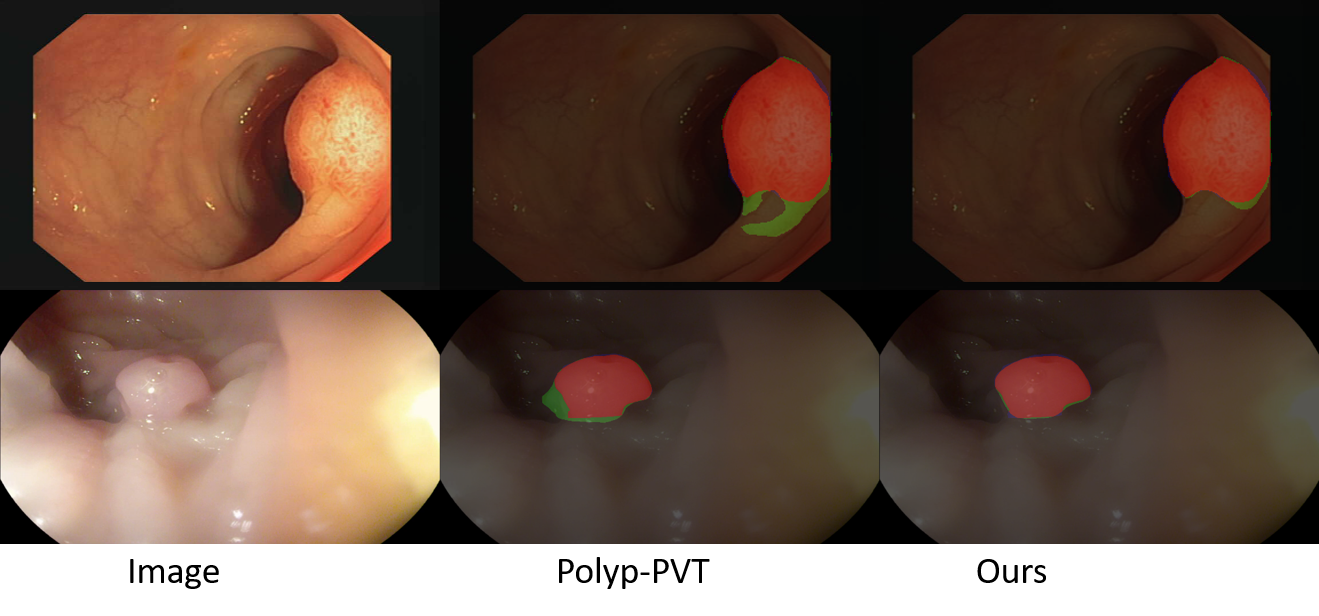}
\caption{Main challenge of polyp segmentation is the object boundary, which is ambiguous due to the color and texture of polyps are very similar to surrounding tissues. The segmentation examples of our model and Poly-PVT\cite{dong2021polyp} with different datasets, which show that our model has better ability on the border. Red indicates correct polyp region. Green is the missed polyp region.}
\label{myfig1}
\end{center}
\end{figure}

Although these methods have substantially improved accuracy and generalization ability compared with traditional methods, locating 
the edges of polyps remains a challenge for them, as shown in Figure.\ref{myfig1}. The color and texture of polyps are markedly similar to surrounding tissues; low contrast provides them with powerful camouflage properties \cite{fan2021concealed} and makes them difficult to identify. Previous studies \cite{chen2018encoder, dong2021polyp, shen2021exploring, li2020gated, shen2021efficient} have explored fusing low-scale boundary and high-scale semantic information to preserve boundary details better. Takikawa et al.  \cite{takikawa2019gated} and Zhen et al.\cite{zhen2020joint} designed a boundary stream and coupled the task of boundary and semantics modeling. PraNet \cite{fan2020pranet} generated a global map as the initial guidance region and used the reverse attention module thereafter to reveal more complete objects. However, these endeavors seldom consider simulating how humans detect polyps with their ambiguous boundaries to the backgrounds.

To address the previously mentioned issues, we have proposed a new deep learning model called the Separated Edge-Guidance Transformer Network (SegT) for polyp segmentation. Our model generates high-quality segmentation maps imitating the human manner and has demonstrated remarkable performance in various challenging scenarios. When people detect potential polyp targets in colonoscopy images, they will first look for the possible polyp region. Thereafter, they will outline the precise edge of the polyp area by comparing the difference between the foreground and background. Inspired by this observation, we propose an effectively separated edge-guidance for polyp segmentation.

The key contributions of our work are as follows:
\begin{itemize}
 \item We propose a novel framework called separated edge-guidance transformer network (SegT) for polyp segmentation, which adapts the pyramid vision transformer (PVT) as encoder rather than the existing CNN-based methods to extract features.

 \item We design a separated edge-guidance (SEG) module, which is composed of two parts: separator (SE) and edge-guidance (EG) blocks. Their purpose is to simulate how humans detect polyp targets. In particular, the SE block is utilized to highlight the object’s edges between an image’s background and foreground. The EG block aims to embed edge information into the feature map, which can significantly address the “ambiguous” problem of edges.

 \item We present a cascade fusion module (CFM), which collects polyps’ semantic and location information from the features through progressive integration to obtain refined segmentation results.
 \end{itemize}

\section{Related work}

\subsection{Polyp segmentation}

\emph{\textbf{Traditional Methods.}}  Computer-aided detection \cite{sanchez2018automatized,  camnet, figueiredo2019polyp} is an effective alternative to manual detection, and hand-engineered methods are widely used in polyp detection. The methods of polyp segmentation schemes are mainly based on low-level features, such as texture and geometric features. In the method proposed by Gonzalez et al. \cite{sanchez2018automatized}, the shape, color, and curvature features of edges are utilized for polyp segmentation. Figueiredo et al. \cite{figueiredo2019polyp} proposed a unified bottom-up and top-down saliency method for polyp detection that considers shape, color, and texture information. However, these methods have a high risk of missed or false detection owing to the high similarity between polyps and the surrounding tissues.

\emph{\textbf{Deep Learning-Based Methods.}}  Owing to the powerful feature expression and analysis capabilities of deep learning models \cite{brandao2018towards, shen2022hsgm, ronneberger2015u, shen2019large, fan2020pranet}, many deep learning-based methods have been proposed for polyp segmentation tasks. Brandao et al. \cite{brandao2018towards} introduced FCN into the polyp region extraction issue by converting the classification neural network. However, this fully convolutional network architecture lacks detailed semantic features, and the segmentation result is not ideal. Encoder–decoder-based models, such as U-Net \cite{ronneberger2015u} and UNet++ \cite{zhou2018unet++}, have recently become important model frameworks in this direction, which have excellent performance. U-Net \cite{ronneberger2015u} introduced incremental up-sampling of feature maps alongside the corresponding scales of low-level feature maps with “skip-connections.” U-Net++ \cite{zhou2018unet++} included additional layers and dense connections, which are used to reduce the gap between low- and high-level features. With the increasing importance of polyp segmentation, attention mechanism \cite{li2022enhancing} has been designed specifically for polyp datasets in recent years. PraNet \cite{fan2020pranet} utilizes a reverse attention module to establish the relationship between region and boundary cues, recovering a clear boundary between a polyp and its surrounding mucosa. However, solely using reverse attention may lead to false detections and introduce unnecessary noise. Inspired by Chen et al. \cite{chen2018reverse}, we adopt the separate attention mechanism, which combines reverse and normal attention to focus on the background and foreground, respectively.

\subsection{Vision Transformer}

Transformer \cite{vaswani2017attention} is a markedly influential deep neural network architecture originally proposed to solve similar problems, such as natural language processing. Originally, the transformer architecture was not well suited for image analysis. To apply transformers to computer vision tasks, Dosovitskiy et al. \cite{dosovitskiy2020image} proposed a vision transformer (ViT), which is the first pure transformer for image classification. ViT splits an image into patches and processes them as consecutive labels. This method substantially reduces the computational cost and enables transformers to efficiently process large-scale images. However, ViT requires large-scale datasets to train effectively and is severely limited when trained on small datasets. This property hinders its usage in such problems as medical segmentation, where the dataset is scarce. 

Recent studies have attempted to enhance ViT in several ways further. DeiT \cite{touvron2021training} introduces a data-efficient training strategy combined with a distillation method, which helps improve performance when training on small datasets. HVT \cite{pan2021scalable} is based on a hierarchical progressive pooling method to compress the sequence length of tokens, reducing redundancy and computation. TNT \cite{han2021transformer} adopts a transformer suitable for fine-grained image tasks to segment the original image patch and perform self-attention mechanism calculations in small units. Simultaneously, global and local features are extracted using external and internal transformers. Previous research has demonstrated that the pyramid structure in convolutional networks is also applicable to transformers and various downstream tasks, such as Swin Transformer \cite{liu2021swin}, PVT \cite{wang2022pvt}, and Segformer \cite{xie2021segformer}. PVT is less computationally intensive than ViT and uses the classic semantic FPN to deploy semantic segmentation tasks.

In medical image segmentation, the TransUNet \cite{chen2021transunet} and TransFuse \cite{zhang2021transfuse} models are developed based on a transformer for polyp segmentation and have achieved good results. TransUNet uses a transformer-based network with a hybrid ViT encoder and an upsampled CNN decoder. Hybrid ViT stacks CNN and transformer together, resulting in high computational costs. TransFuse solves this problem by using a parallel architecture. Both models use the attention gate mechanism \cite{schlemper2019attention} and the so-called BiFusion module. These components make the network architecture large and highly complex. To efficiently train models on medical images, Poly-PVT \cite{dong2021polyp} introduces a similarity aggregation module based on the graph convolutional domain \cite{lu2019graph}.

\subsection{Image Edge Segmentation}

Locating pixels on the border is considerably  difficult, as demonstrated by many previous methods. To address the issues, various edge-aware models (or say boundary-aware models) have been developed to highlight these hard pixels.

Learning edge information has shown excellent performance in many image segmentation tasks in recent years. In early studies on FCN-based semantic segmentation, Bertasius et al. \cite{bertasius2016semantic} and Chen et al. \cite{chen2017deeplab} used boundaries for post-starting to refine the results at the end of the network. SFANet \cite{fang2019selective} applies region boundary constraints to supervise polyp learning. To compensate for missing object parts, Chen et al. \cite{chen2020reverse} and Fan et al. \cite{fan2020pranet} utilized reverse attention blocks to learn missing parts and details. However, using only edge information as shape constraints or reverse attention may lead to incorrect detection or introduce unnecessary noise. Several recent approaches have explicitly parallelized boundary detection as an independent subtask with semantic segmentation to achieve cleaner results. Ma et al. \cite{ma2021boundary} explicitly exploited boundary information for context aggregation, further enhancing the semantic representation of the model. Kim et al. \cite{kim2021devil} went a step further than BlendMask \cite{chen2020blendmask} and explored base mask representations and boundary information for instance-specific features.

Although the preceding methods can improve performance, they only use boundary information as supplementary clues to effectively refine the target region segmentation. These methods minimally exploit the complementary relationship between regional and boundary features. Compared with the methods above, our proposed method can mine the deep information of the foreground and background and combine the boundary information to enhance the features at the junction of the foreground and background, thereby improving the segmentation performance of the polyp targets.

\begin{figure}[t]
\begin{center}
\includegraphics[scale=0.5]{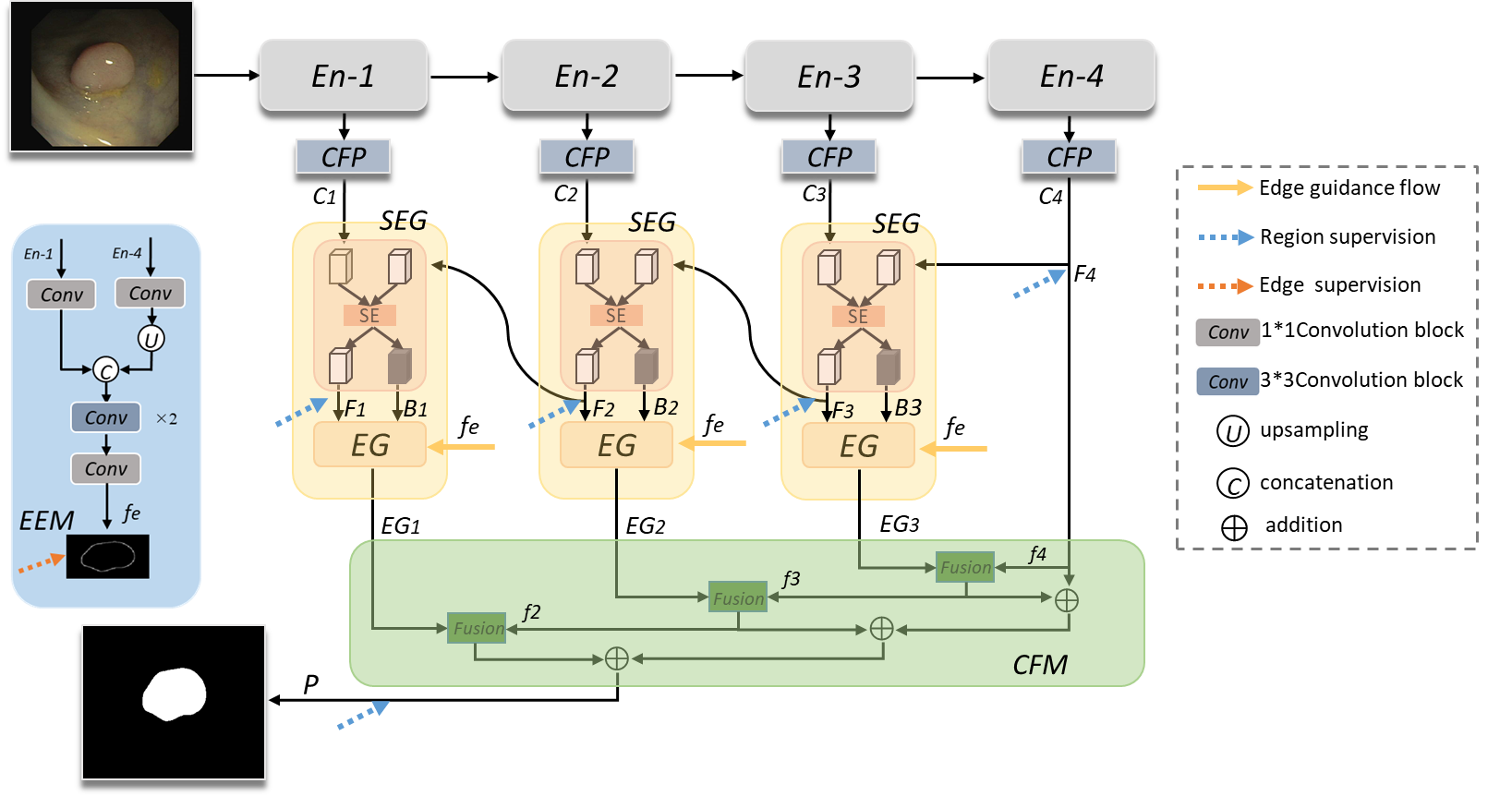}
\caption{Overview of the proposed SegT, which consists of three key components: edge extractor (EEM) module , separated edge-guidance (SEG) module, and cascade fusion module (CFM).}
\label{myfig2}
\end{center}
\end{figure}

\section{Methodology}

The full architecture network is shown in Figure \ref{myfig2}. For the input $I\in R^{W \times H \times 3}$  , where $W$ and $H$ denote the width and height of an image, we use pyramid vision Transformer (PVT) as our backbone to extract the multi-level features $En_i,i\in \{1,2,3,4\}$. First, we input $En_1, En_2, En_3$ and $En_4$ into the channel-wise feature pyramid (CFP)\cite{lou2021cfpnet} to determine the features of different receptive fields. Second we utilize separated edge-guidance module (SEG) to refine the feature maps. The SEG module consists of separator blocks (SE) and edge-guidance blocks (EG). The SE blocks contain the normal and reverse attention streams to focus on the foreground and background. The foreground maps $F_i, i \in \{1,2,3,4\}$ are supervised by the ground truth. Furthermore, we utilize an effective edge extractor module (EEM) to obtain the edge map, which is exploited in the EG blocks. Thereafter, we obtain the edge refined maps marked as $EG_i,i \in {1,2,3,4}$ and feed them to the cascade fusion module (CFM) to fuse refined feature maps, leading to a final feature map $P$. We choose the sum of $EG_1$ and $P$ as the final output in the inference stage.

\subsection{Transformer encoder}

Some recent works\cite{xie2021segformer, bhojanapalli2021understanding} report that vision transformer\cite{wang2022pvt, wang2021pyramid} have stronger performance and robustness to input disturbances such as noise than CNNs\cite{he2016deep, simonyan2014very}. Inspired by this, we choose a vision transformer as our backbone network to extract features for polyp segmentation. Compared with\cite{dosovitskiy2020image, liu2021swin}, the PVT\cite{wang2022pvt} is a pyramid architecture whose spatial-reduction attention can reduce the computing resource consumption. To the segmentation task, we design a polyp segmentation head on top of four multi-level feature maps (i.e. $En_1, En_2, En_3, En_4$). Among these feature maps, $En_1$ gives detail low level appearance information, $En_2, En_3, En_4$ provides high-level feature.

\subsection{Edge extractor module}

Object detection can benefit from a good edge prior in segmentation and localization\cite{zhao2019egnet}. Even though low-level features contain rich edge details, they introduce non-object edges information. To easily to explore edge features associated with the polyp area, high-level semantic or location information is required. As illustrated in Figure \ref{myfig2}, we combine the high-level feature ($En_4$) and low-level feature ($En_1$) in this module to model the object-related edge information. First, the channels of $En_1$ and $En_4$ are separately changed to 32 and 256, respectively, using two 1×1 convolution layers. Second, the feature $En_1'$ and up-sampled feature $En_4'$ are combined using a concatenation technique. Lastly, we generate the edge feature $f_e$ using two 3×3 convolution layers and one 1×1 convolution layer. The produced edge map and its edge ground-truth label can be measured using the binary cross-entropy loss function, which is given as follows:
\begin{equation}
L_{e d g e}=-\sum_i\left[E_i^{g t} \log \left(E M_i\right) +\left(1-E_i^{g t}\right) \log \left(1-E M_i\right)\right], \label{eq1}\end{equation}
where $EM_i$ denotes the produced edge map of the \emph{i}-th image after the upsampling operator of the edge feature $f_e$, and $E_i^{g t}$ denotes the edge ground-truth map. In our model, the Canny edge detection method is used to extract $E_i^{g t}$. Moreover, our EEM can provide edge-enhanced representation $f_e$ to guide the detection in the separated edge-guidance module. In addition, $f_e$ is cascaded to multiple supervisions to enhance the ability of feature representations.

\begin{figure}[t]
\begin{center}
\includegraphics[scale=0.5]{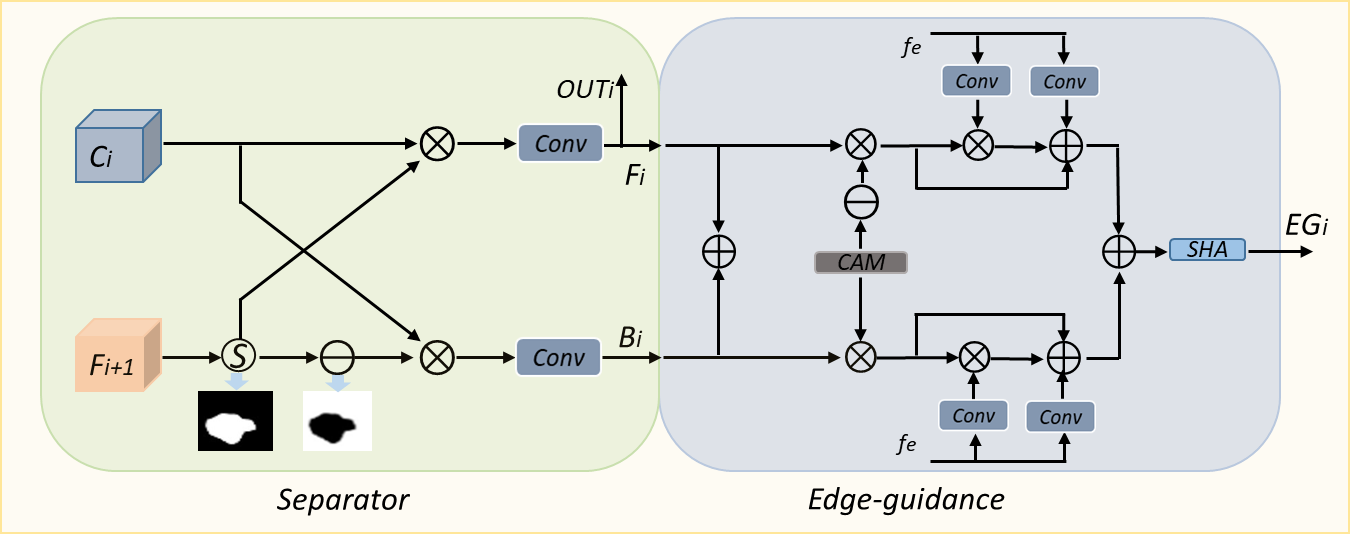}
\caption{ Illustration of the proposed separated edge-guidance (SEG) module, consisting of two key components: separator (SE) block and edge-guidance (EG) block.}
\label{myfig3}
\end{center}
\end{figure}

\subsection{Separated edge-guidance module}

As shown in Figure \ref{myfig3}, the original input images have blurred boundaries that disguise areas of polyps that are difficult to segment. To address these issues, a separated edge-guidance (SEG) module is proposed. The module integrates the SE and EG blocks. The SE block contains forward and reverse streams to focus on the foreground and background, respectively. The EG block integrates edge information from EEM into the feature space to enhance the sensitivity of the model to the edge.

\emph{\textbf{Separator.}} When delineating the polyp area in colonoscopic images, information at the boundary between the foreground and background is an important cue. The human vision system is able to perceive border information effectively because of the fusion of information from the background and interior of the object. Inspired by \cite{chen2018reverse}, we adopt the SE block, which contains two steams to focus on the foreground and background. In the first stream, we erase the internal details of objects to focus on the background. Meanwhile, the internal information of the object is recovered in the second stream to focus on the foreground. The operation mechanism of the separator is to highlight the boundary through the synergy between the foreground and background information. The separator can be written as follows:
\begin{equation}
F_i=O u t_i=\operatorname{Conv}\left(C_i \otimes \operatorname{expand}\left(\sigma\left(F_{i+1}\right)\right)\right), \label{eq2}\end{equation}
\begin{equation}
B_i=\operatorname{Conv} \left(C_i \otimes \operatorname{expand}\left(1-\sigma\left(F_{i+1}\right)\right)\right), \label{eq3}\end{equation}
where $C_i$ denotes the \emph{i}-th layer of the feature map produced by the CFP module \cite{lou2021cfpnet}. The foreground map in the \emph{i}-th layer $F_i$ is the result of upsampling on the coarse map of the $\left( i+1 \right)$-th layer, written as $ 1 - \sigma \left( F_{i+1}\right)$, where $\sigma$ denotes the sigmoid function, $\operatorname{Conv}$ is a 1 × 1 convolution, $\otimes$ indicates the multiplication operation, and $\operatorname{expand \left( \right)}$ aims to expand the channel of maps similar to $C_i$. The background map is the foreground map substracted from 1, which is defined as $1-\sigma\left( F_{i+1}\right)$. $Out_i$ is the coarse output map of the \emph{i}-th layer, which is supervised by the ground truth map.

\emph{\textbf{Edge-guidance.}} We use $F_i$ and $B_i$ as the input to Channel Attention Module (CAM) \cite{dai2021attentional}, which is beneficial in representing different scales of features in a more general way. Moreover, the attention that obtains the weight of the feature maps on a global and local scale can be written as $ \operatorname{W} \left(F_i+B_i\right)$. After the attention module, we add an edge guider to enhance the model’s ability to understand the edges, so they can be more prominent after the two streams are merged. In particular, we integrate the information of the two streams by simple addition. The mechanism of the edge guider module is similar to the conditional normalization module with a prior knowledge of edge map. We consider edge prediction as our condition, and such a module embeds the spatial information into  feature maps $F_i$ and $B_i$, which allows the feature map to learn better edge features. The formula of the operation is defined as follows:
\begin{equation}
\begin{split}
EGF_i=B N\left(W\left(F_i+B_i\right) \otimes F_i\right)\otimes \operatorname{Conv}_{3 \times 3}\left(f_e\right)
\oplus \operatorname{Conv}_{3 \times 3}\left(f_e\right),    
\end{split}
 \label{eq4}\end{equation}
\begin{equation}
\begin{split}
 EGB_i=B N\left(\left(1-W\left(F_i+B_i\right)\right) \otimes B_i\right) \otimes \operatorname{Conv}_{3 \times 3}\left(f_e\right)
 \oplus \operatorname{Conv}_{3 \times 3}\left(f_e\right),   
\end{split}
 \label{eq5}\end{equation}
 \begin{equation}
EG_i=EGF_i \oplus EGB_i , i=1,2,3,4,
 \label{eq6}\end{equation}
where $EGF_i$ and $EGB_i$  are the output results of the foreground and background streams respectively, $EG_i$ is the output of the SEG module, $f_e$ is the edge feature map, \emph{BN} denotes batch normalization, and $Conv_{3 \times 3}$ denotes a 3×3 convolutional layer to encode information on the edge map and enlarge the channel to the same as coarse feature maps. Thereafter, the shuffle attention module (SAM) \cite{zhang2021sa} is utilized to make the model focus on the informative channels.

\subsection{Cascade fusion module}

A multi-level feature fusion strategy was applied and verified to improve segmentation performance. The feature fusion has immense influence on the quality of the segmentation result, so we design a cascade fusion module (CFM) to achieve more effective output feature fusion. CFM obtains four edge-guided maps of the first round predictions marked as \{ $EG_i,i=1,…,4$ \}. Each lower-level feature map is aggregated with the result of the fusion process. The process can be summarized as $ \operatorname{Fusion} \left( f_i, EG_i \right) = \operatorname{Concat} \left( f_i \otimes EG_i, EG_i \right)$. The four levels of fusion feature stacks are shown in Figure \ref{myfig2}. The four levels of fusion feature \{ $f_i,i=1,…,4$ \} are computed using Equation \ref{eq7}, and the final output is computed using $\sum_i f_i$ .

 \begin{equation}
\left\{\begin{array}{c}f_4=C_4 \\ f_3=F u \operatorname{sion}\left(f_4, E G_3\right) \\ f_2=\operatorname{Fusion}\left(f_3, E G_2\right) \\ f_1=\operatorname{Fusion}\left(f_2, E G_1\right)\end{array}\right.,
 \label{eq7}\end{equation}

\subsection{Loss function}

Binary cross-entropy loss is widely used in many polyp segmentation tasks. However, it has clear shortcoming that will lead to poor performances when the number of foreground pixels considerably less than that of background pixels. Inspired by\cite{dong2021accurate}, we combine the two loss functions as the total loss for supervision with the following formula: 
 \begin{equation}
L_t=L_{w b c e}+L_{w I O U},
 \label{eq8}\end{equation}
where $L_{wIOU}$ and $L_{wbce}$ denote the weighted IoU loss and BCE loss for global and local restrictions, respectively. Note that $L_{wIOU}$ can increase the weights of hard pixels to highlight their importance, and $L_{wbce}$ focuses more on hard pixels rather than treating all pixels equally. Moreover, our model includes six supervised outputs, including four foreground maps $ \left( F_1,F_2,F_3,F_4 \right)$, one feature fusion map $P$, and one edge map $f_e$. Each map (i.e.,$F_1,F_2,F_3,F_4,P$ ) is up-sampled to have the same size as the ground-truth map (i.e., $G$). Thus, the final total loss function can be represented as follows:
\begin{equation}
L_{\text {total }}=\sum_{i=1}^4 L_t\left(F_i, G\right)+L_t(P, G)+L_{\text {edge }}.
\label{eq8}\end{equation}

\section{Experiments}

\subsection{Datasets and evaluation metrics}

Following PraNet \cite{fan2020pranet}, we conduct experiments on five polyp segmentation datasets (ETIS \cite{vazquez2017benchmark}, CVCClinicDB (ClinicDB) \cite{silva2014toward}, CVC-ColonDB (ColonDB) \cite{bernal2015wm}, EndoScene-CVC300 (EndoScene) \cite{tajbakhsh2015automated}, Kvasir-SEG (Kvasir) \cite{jha2020kvasir}). Our training set contains 900 randomly selected images in Kvasir and 550 selected images in CVC-ClinicDB, while the remaining 100 pieces of Kvasir and 62 pieces of CVC-ClinicDB are used as test sets. Test on the out-of-distribution datasets includes ColonDB with 380 images, EndoScene with 60 images, and ETIS with 196 images (unseen data). Three widely used metrics, namely, mean Dice (mDice), mean IoU (mIoU) and mean absolute error (MAE), are used to evaluate the model performances.

\subsection{Implementation Details}

Our method is implemented based on the PyTorch framework and runs on an NVIDIA GeForce RTX 3090 GPU. Considering the differences in the sizes of each polyp image, the input image is simply resized to 352 × 352, and we adopt a multi-scale training strategy thereafter \cite{fan2020pranet, shen2021competitive, huang2021hardnet}. The network is trained end-to-end by an AdamW \cite{loshchilov2017decoupled} optimizer. The learning rate is set to 1e-4, and the weight decay is adjusted to 1e-4 as well. The batch size is set at 16.

\begin{table*}[t]
\centering
\caption{Quantitative comparison of different methods on the Kvasir and ClinicDB(seen datasets) to validate our model’s learning ability. ↑ denotes higher the better and ↓ denotes lower the better.}

\scalebox{0.8}{
\begin{tabular}{ccccccccc}
\hline
                          &                        &                        & \multicolumn{3}{c}{Kvasir}                                             & \multicolumn{3}{c}{ClinicDB}                                                  \\ \cline{4-9} 
\multirow{-2}{*}{Methods} & \multirow{-2}{*}{Pub.} & \multirow{-2}{*}{Type} & mDice↑ & mIoU↑ & MAE↓  & mDice↑         & mIoU↑ & MAE↓ \\ \hline
U-Net\cite{ronneberger2015u}                     & MICCAI'15              & CNN                    & 0.818                          & 0.746                         & 0.055 & 0.823          & 0.755                         & 0.019                        \\
UNet++\cite{zhou2018unet++}                    & TMI'19                 & CNN                    & 0.821                          & 0.743                         & 0.048 & 0.794          & 0.729                         & 0.022                        \\
PraNet\cite{fan2020pranet}                    & MICCAI'20              & CNN                    & 0.898                          & 0.840                         & 0.030 & 0.899          & 0.849                         & 0.009                        \\
CaraNet\cite{lou2022caranet}                   & JMI'23                 & CNN                    & 0.918                          & 0.865                         & 0.023 & 0.936          & 0.887                         & 0.007                        \\
TransUNet\cite{chen2021transunet}                 & arXiv'21               & Transformer            & 0.913                          & 0.857                         & 0.028 & 0.935          & 0.887                         & 0.008                        \\
TransFuse\cite{zhang2021transfuse}                 & MICCAI'21              & Transformer+CNN        & 0.920                          & 0.870                         & 0.023 & \textbf{0.942} & 0.897                         & 0.007                        \\
Polyp-PVT\cite{dong2021polyp}                 & CAAI   AIR'23          & Transformer            & 0.917                          & 0.864                         & 0.023 & 0.937          & 0.889                         & 0.006                        \\
SegT (Ours)               & -                      & Transformer            & \textbf{0.927}                 & \textbf{0.880}                & \textbf{0.023}                & 0.940                                  & \textbf{0.897}                & \textbf{0.006}               \\ \hline
\end{tabular}}
\label{table1}
\end{table*}

\begin{table*}[t]
\centering
\caption{Quantitative comparison of different methods on the ColonDB, ETIS and EndoScene datasets (unseen datasets) to validate our model’s generalization capability. ↑ denotes higher the better and ↓ denotes lower the better.}
\scalebox{0.75}{
\begin{tabular}{cccccccccccc}
\hline
                          &                        &                        & \multicolumn{3}{c}{ColonDB}                                            & \multicolumn{3}{c}{ETIS}                                               & \multicolumn{3}{c}{EndoScene}                                                 \\ \cline{4-12} 
\multirow{-2}{*}{Methods} & \multirow{-2}{*}{Pub.} & \multirow{-2}{*}{Type} & mDice↑ & mIoU↑ & MAE↓  & mDice↑ & mIoU↑ & MAE↓  & mDice↑         & mIoU↑ & MAE↓ \\ \hline
U-Net \cite{ronneberger2015u}                    & MICCAI'15              & CNN                    & 0.512                          & 0.444                         & 0.061 & 0.398  & 0.335                         & 0.036 & 0.710          & 0.627                         & 0.022                        \\
UNet++ \cite{zhou2018unet++}                    & TMI'19                 & CNN                    & 0.483                          & 0.410                         & 0.064 & 0.401  & 0.344                         & 0.035 & 0.707          & 0.624                         & 0.018                        \\
PraNet  \cite{fan2020pranet}                  & MICCAI'20              & CNN                    & 0.712                          & 0.640                         & 0.043 & 0.628  & 0.567                         & 0.031 & 0.851          & 0.797                         & 0.010                        \\
CaraNet \cite{lou2022caranet}                  & JMI'23                 & CNN                    & 0.773                          & 0.689                         & 0.042 & 0.747  & 0.672                         & 0.017 & \textbf{0.903} & \textbf{0.838}                & \textbf{0.007}               \\
TransUNet \cite{chen2021transunet}                & arXiv'21               & Transformer            & 0.781                          & 0.699                         & 0.036 & 0.731  & 0.824                         & 0.021 & 0.893          & 0.660                         & 0.009                        \\
TransFuse  \cite{zhang2021transfuse}               & MICCAI'21              & Transformer+CNN        & 0.781                          & 0.706                         & 0.035 & 0.737  & 0.826                         & 0.020 & 0.894          & 0.654                         & 0.009                        \\
Polyp-PVT \cite{dong2021polyp}                & CAAI   AIR'23          & Transformer            & 0.808                          & 0.727                         & 0.031 & 0.787  & 0.706                         & 0.013 & 0.900          & 0.833                         & 0.007                        \\
SegT (Ours)               & -                      & Transformer            & \textbf{0.814}                 & \textbf{0.732}                & \textbf{0.026}                & \textbf{0.810}                 & \textbf{0.732}                & \textbf{0.013}                & 0.895                                  & 0.828                         & 0.008                        \\ \hline
\end{tabular}}
\label{table2}
\end{table*}

\subsection{Result}

We first evaluate our proposed SegT model for its segmentation performance on the seen datasets. As summarized in Table  \ref{table1}, our model is compared to four recently published CNN-based neural networks: U-Net \cite{ronneberger2015u}, UNet++ \cite{zhou2018unet++}, PraNet \cite{fan2020pranet}, and CaraNet \cite{lou2022caranet}. Note that our proposed model outperforms other models on the seen datasets, as shown in Table \ref{table1}. In the first two rows of the table, we compared two classic medical image segmentation networks (the U-Net and the U-Net++). The SegT network achieves over 10\% gains in mDice and mIoU on the Kvasir datasets and ClinicDB. In the 3rd and 4th rows, we compared state-of-the-art models of the polyp segmentation task. Our proposed model outperforms the two models in mDice, mIoU, and MAE on the seen datasets. In Table \ref{table1}, we also report the results of three transformer-based methods (TransUNet \cite{chen2021transunet}, TransFuse \cite{zhang2021transfuse}, and Polyp-PVT \cite{dong2021polyp}) with our proposed framework. Although TransFuse is close to the performance of our proposed model in the ClinicDB dataset, our model is more stable in terms of overall performance. Furthermore, without considering model complexity, our model has a 1\% improvement in the mIoU metric in the challenging Kvasir dataset. When compared to the other two types of transformer-based models, the advantage of the metric performance is more obvious.

We further evaluate the generalization capability of our model on unseen datasets (i.e., ETIS, ColonDB, EndoScene). Table \ref{table2} shows that our model outperforms the existing medical segmentation baselines on the unseen datasets. Concretely, performance gains over the best contender built on a CNN-based backbone network (i.e., CaraNet) are (4.1\%, 4.3\%, 0.016) for metrics (mDice, mIoU, MAE) on the ColonDB dataset, and (6.3\%, 6\%, 0.004) for metrics (mDice, mIoU, MAE) on ETIS dataset. Besides, when compared with transformer-based backbone networks on the challenging ETIS datasets, our SegT surpasses the best competing method (i.e., Polyp-PVT) by 2.3\% and 2.5\% for the mDice and mIoU metrics, respectively. However, our evaluation results on the EndoScene dataset show that our method doesn't demonstrate a significant performance advantage compared to other approaches. This outcome is mainly due to the fact that the test dataset consists of only 60 images, which hardly draws definitive conclusions about the superiority of the different methods.

\begin{figure}[t]
\begin{center}
\includegraphics[scale=0.5]{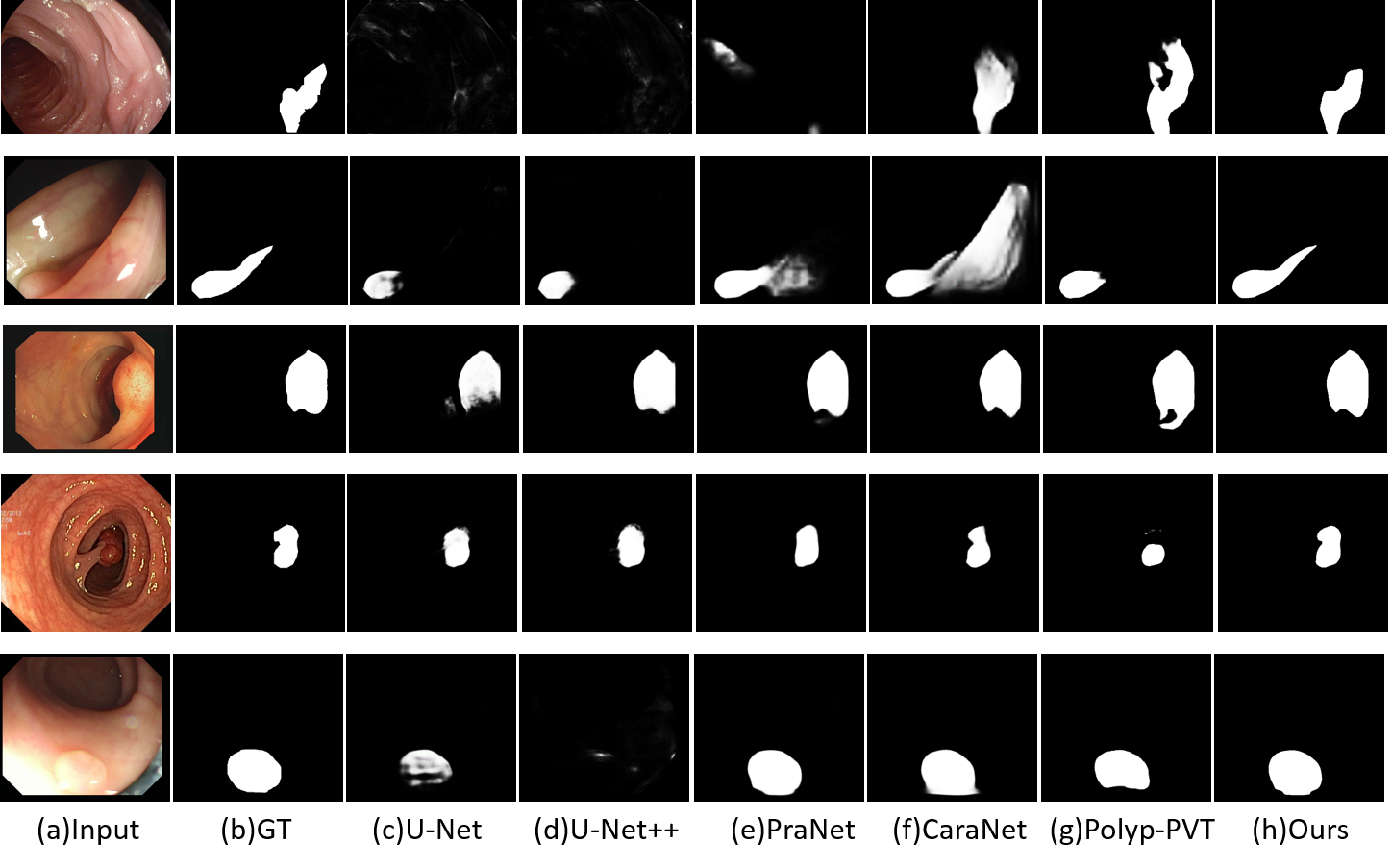}
\caption{Qualitative results of different methods. (a) Inputs images, (b) GT, which means the ground truths, (h) semantic segmentation maps produced by our method, (c) U-Net \cite{ronneberger2015u} , (d) U-Net++ \cite{zhou2018unet++}, (e) PraNet \cite{fan2020pranet}, (f) CaraNet \cite{lou2022caranet}, (g) Polyp-PVT \cite{dong2021polyp}.}
\label{myfig4}
\end{center}
\end{figure}

To intuitively demonstrate the prominent performance of the SegT, several representative prediction maps of the SegT and other state-of-the-art models are shown in Figure \ref{myfig4}. Specifically, the 1st shows the results of examples with low-contrast backgrounds. As we can see, while most competing methods cannot identify the boundary area, our SegT almost correctly segments all the polyp regions. The 2nd row is an example of the occluded polyp. As we can observe, our SegT is capable of producing accurate results, while other methods tend to generate results with poor accuracy. The 3rd and 4th rows are examples with relatively large targets and small targets, respectively. As we can observe, our SegT correctly identifies all the targets. Other methods tend to miss several details of the boundaries. The 5th row is an example of brightness interference. Our SegT not only accurately segments the target but also eliminates the salient distraction. In summary, SegT is capable of producing high-quality prediction maps under various challenging scenarios. It is worth noting that in these examples, the texture of these polyp objects is almost identical to that of the surrounding environment, which can well prove that the SegT is effective in locating the targets by leveraging the edge cues.

\subsection{Ablation Study}

We describe in detail the effectiveness of each component on the overall model. The training, testing, and hyper-parameter settings are the same as mentioned in Sec.B. We evaluate module effectiveness by removing components from the complete SegT on three datasets, and we choose mDice, mIoU and MAE for evaluation. In order to better explain the relationship between models, we labeled different experimental models as a to e. Model a is composed of the backbone network PVTv2 and Channel-wise Feature Pyramid (CFP) module; Model b adds the Separator block (SE) based on Model a; Model c adds separated edge-guidance module (SEG) on the Model a; Model d is the final model without Cascade fusion module, and Model e is our final model. We evaluate the seven models on three benchmark datasets. Quantitative experimental results are shown in Table \ref{table3}.

\emph{\textbf{Effectiveness of SEG.}} By comparing Model a with Model b, we observe that Model b outperforms Model a in terms of all the evaluation metrics. It means by adding the separator block, our model can perform better. The apparent improvement in the evaluation metrics shows that the separator can highlight the boundaries of objects by focusing on the foreground and background information separately, thereby improving the accuracy of polyp segmentation. In order to validate the effectiveness of the Edge-guidance, we compare the results of Model b and Model c. After adding the Edge-guidance (EG), the performance of our Model c increases compared with Model b. Moreover, we further investigate the contribution of the SEG by removing it from the overall model, which is labeled as Model d, the performance without the SEG drops sharply on all three datasets. Compared with Model d, Model e and Model f shows an improvement, which demonstrates the two block in the SEG module work effectively. Since the separator between the foreground and the background, that is, the boundary of the polyp area contains fewer pixels, we need to exploit the Edge-guidance to embed additional edge information into the feature to strengthen the model’s understanding of boundary. With the help of Edge-guidance, the predicted result can maintain a clear edge structure of the object.

\emph{\textbf{Effectiveness of CFM.}} Similarly, we test the effectiveness of the CFM module by removing it from the overall model and replacing it with an element-wise addition operation, which is called Model c. Compared with SegT, the performance of the Model c drops on all three datasets by a large margin. The performance degradation of the model demonstrates that the CFM is helpful in effectively integrating refined feature information at every stage. By comparing Model a with Model d, the baseline model with the CFM module also can perform better in most of the evaluation metrics.

\begin{figure}[t]
\begin{center}
\includegraphics[scale=0.5]{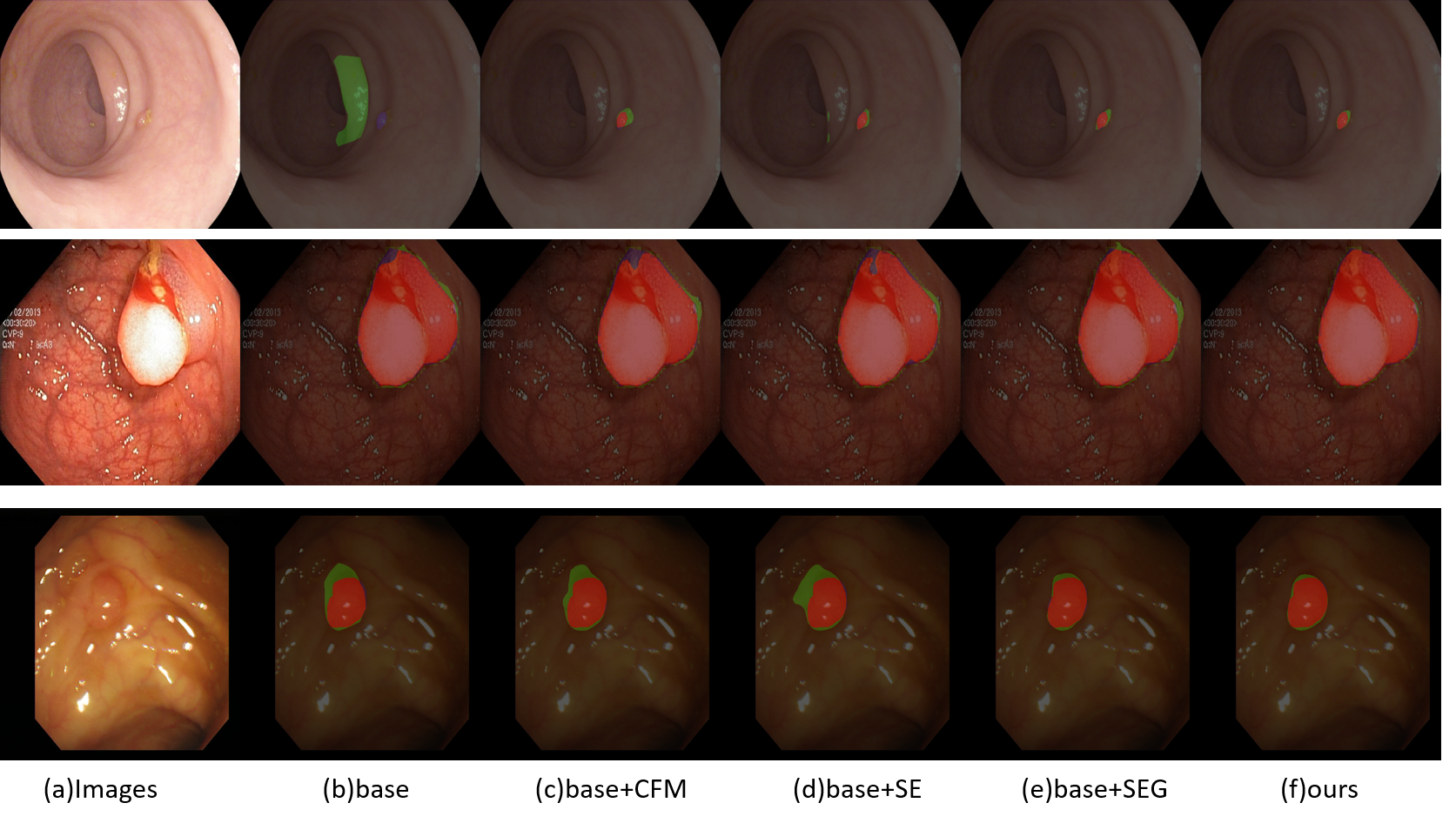}
\caption{ Effectiveness of each component. Green indicates regions that are not detected. Red represents regions that are accurately detected.}
\label{myfig5}
\end{center}
\end{figure}

The visual results are given in Figure \ref{myfig5}. Green and red indicate regions that are not detected and accurately detected, respectively. Evidently, our designed module can obtain significant results in the edge detection of small and large target regions. We observe that the SEG module facilitates the fine-grained ambiguous boundaries, and the CFM module significantly improves the accuracies of object detection and target object location.

\begin{table*}[]
\centering
\caption{Ablation study for SegT on the Kvasir, ETIS and ColonDB datasets. ↑ denotes higher the better and ↓ denotes lower the better.}
\scalebox{0.75}{
\begin{tabular}{l c c c c c c c c c }
\hline
                          & \multicolumn{3}{c}{Kvasir(seen)} & \multicolumn{3}{c}{ETIS(unseen)} & \multicolumn{3}{c}{ColonDB(unseen)} \\ \cline{2-10} 
\multirow{-2}{*}{Methods} & mDice↑            & mIoU↑             & MAE↓             & mDice↑            & mIoU↑             & MAE↓             & mDice↑             & mIoU↑              & MAE↓              \\ \hline
a. baseline                                       & 0.910             & 0.859             & 0.030            & 0.759             & 0.688             & 0.017            & 0.796              & 0.707              & 0.031             \\
b. baseline +  SE                                 & 0.914             & 0.856             & 0.033            & 0.767             & 0.707             & 0.018            & 0.799              & 0.721              & 0.031             \\
c.  baseline +    SEG (SE + EG)                   & 0.919             & 0.869             & 0.028            & 0.795             & 0.714             & 0.016            & 0.810              & 0.727              & 0.030             \\
d.  baseline + CFM                                & 0.913             & 0.855             & 0.034            & 0.764             & 0.701             & 0.018            & 0.792              & 0.721              & 0.032             \\
e.  w/o SE                                        & 0.916             & 0.865             & 0.030            & 0.779             & 0.701             & 0.019            & 0.794              & 0.710              & 0.032             \\
f.   w/o EG                                       & 0.914             & 0.861             & 0.031            & 0.777             & 0.703             & 0.019            & 0.798              & 0.709              & 0.031             \\
\textbf{g.  SEG+ CFM (Ours)}                      & \textbf{0.927}    & \textbf{0.880}    & \textbf{0.023}   & \textbf{0.810}    & \textbf{0.732}    & \textbf{0.013}   & \textbf{0.814}     & \textbf{0.732}     & \textbf{0.026}    \\ \hline
\end{tabular}
}
\label{table3}
\end{table*}

\section{Conclusion}

We proposed a new image polyp segmentation framework called SegT. SegT is inspired by the habit of observing objects with blurred boundaries by finding the foreground and background, the outline of the object can be depicted. Therefore, this research argues that the boundary information will enhance the ability of polyp segmentation. On the bases of the preceding observations, we first utilize a PVT backbone as an encoder to explicitly extract more powerful and robust features. Thereafter, we propose a SEG module composed of two blocks (i.e., SE and EG blocks). The SE block is used to separate two streams: one stream focuses on the foreground but disregards the background, while the other focuses on the background and erases the foreground. After each stream, edge information is embedded into the features using the EG block, and the two streams are fused to enhance the ability of the model to detect object boundaries. Lastly, CFM is used to obtain more accurate features. Extensive experiments show that SegT consistently outperforms on five challenging datasets without any pre-/post-processing.

\textbf{Futher Work.} Although the SegT model provides a powerful and effective solution for the polyp segmentation task, some limitations still deserve further exploration. First, in the current work, the boundary information is collected explicitly to guide the foreground and background to find the subtle differences between the two to depict the boundary of polyps. In contrast, for humans, the information extraction and integration process should be implicitly included in the knowledge-learning process. Moreover, this design brings additional inference costs. In future work, we will simplify the inference structure further to make it more consistent with the actual human decision-making process. In addition, the backbone of the SegT model was pre-trained on ImageNet, where most natural images differ from medical images. In future work, we will use pre-training that is more suitable for medical image segmentation and adapt the model structure to use it for 3-D medical imaging segmentation.

%%%%%%%%%%%%%%%%%%%%%%%%%%%%%%%%%%%%%%%%%%%%%%%%%%%%%%
%          AI TOOLS, USE AND LOCATION
%%%%%%%%%%%%%%%%%%%%%%%%%%%%%%%%%%%%%%%%%%%%%%%%%%%%%%
%We follow COPE's guidelines and policies regarding the use of Artificial Intelligence (AI) tools. COPE Policy on AI tools can be found at https://publicationethics.org/cope-position-statements/ai-author.

%Authors using AI tools in the writing of a manuscript, production of images or graphical elements of the paper, or in the collection and analysis of data, must be transparent in disclosing in this section how the AI tool was used and which tool was used. Authors are fully responsible for the content of their manuscript, even those parts produced by an AI tool, and are thus liable for any breach of publication ethics. - COPE

%Disclosure instructions

%If there is nothing to disclose, there is no need to add a declaration, otherwise please declare.

%\section*{Use of AI tools declaration}
%The author(s) declare(s) they have used Artificial Intelligence (AI) tools in the creation of this article.
%AI tools used:
%How were the AI tools used?
%Where in the article is the information located?

\section*{Use of AI tools declaration}
The authors declare they have not used Artificial Intelligence (AI) tools in the creation of this article.

\section*{Acknowledgments}
We would like to thank Professor Zhaowei Wang from the Shaoxing Hospital and our colleague Ms Xue Cheng for their support. This research was also funded by the university-level key scientific research platform program of Shaoxing University.

\section*{Conflict of interest}

The authors declare there is no conflict of interest.

\bibliographystyle{ieeetr}
\bibliography{ref}
% \begin{thebibliography}{99}

% \end{thebibliography}

\end{document}